\documentclass[preprint,12pt]{elsarticle}
\usepackage{framed,multirow}
\usepackage{subcaption}
\usepackage{multirow}
\usepackage{ragged2e}
\usepackage{amssymb}
\usepackage{latexsym}
\usepackage{makecell}
 
\usepackage{url}
\usepackage{xcolor}
\usepackage[normalem]{ulem}
\definecolor{newcolor}{rgb}{.8,.349,.1}

\makeatletter
\newif\ifemailshortauthor
\makeatother

\emailshortauthortrue

\journal{Pattern Recognition Letters}

\begin{document}

\thispagestyle{empty}

\clearpage

\setcounter{page}{1}

\begin{frontmatter}

\title{Measuring student behavioral engagement using histogram of actions\tnoteref{tn}}
\tnotetext[tn]{This research was funded by NSF Award Number 1900456.}
\author[1]{Ahmed Abdelkawy}\corref{correspondingAuthor}
\author[1]{Aly Farag} \author[1]{Islam Alkabbany} \author[1]{Asem Ali} \author[1]{Chris Foreman} \author[1]{ Thomas Tretter}\author[2]{Nicholas Hindy }
\cortext[correspondingAuthor]{Corresponding author.
\newline \RaggedRight Email Addresses: a0nady01@louisville.edu (A. Abdelkawy), aly.farag@louisville.edu (A. Farag), islam.alkabbany@louisville.edu (I. Alkabbany), asem.ali@louisville.edu (A. Ali), tom.tretter@louisville.edu (T. Tretter), jcfore01@louisville.edu (C. Foreman), hindync@cofc.edu (N. Hindy).
}
  


\affiliation[1]{organization={University of Louisville},
                city={Louisville}, 
                state={KY},
                country={USA}
                }

\affiliation[2]{organization={College of Charleston},
                city={Charleston}, 
                state={SC},
                country={USA}
                }


\begin{abstract}
In this work, we propose a novel method for assessing students' behavioral engagement by representing student's actions and their frequencies over an arbitrary time interval as a histogram of actions. This histogram and the student's gaze are utilized as input to a classifier that determines whether the student is engaged or not. For action recognition, we use students' skeletons to model their postures and upper body movements. To learn the dynamics of a student's upper body, a 3D-CNN model is developed. 
  The trained 3D-CNN model recognizes actions within every 2-minute video segment then these actions are used to build the histogram of actions. 
  To evaluate the proposed framework, we build a dataset consisting of 1414 video segments annotated with 13 actions and 963 2-minute video segments annotated with two engagement levels. Experimental results indicate that student actions can be recognized with top-1 accuracy $86.32\%$ and the proposed framework can capture the average engagement of the class with a 90\% F1-score. 
\end{abstract}


\end{frontmatter}


\section{Introduction}
\label{sec:intro}

Active participation and engagement of students in the classroom is a critical factor for successful learning outcomes \cite{farag2021toward}. Student engagement can be measured through observable behavioral and emotional components that are predictors of cognitive engagement \cite{fredricks2004school}. Behavioral engagement includes self-directed actions taken by students to gain access to the curriculum, such as studying and doing homework, hand and body movements while observing lectures, as well as participating cooperatively in classroom activities \cite{andolfi2017opening}. Emotional engagement, on the other hand, relates to how students feel about their learning experience, environment, instructors, and classmates \cite{skinner1993motivation}. This includes emotions such as excitement or happiness about learning, boredom or disinterest in the material, or frustration and struggle to understand. Lastly, cognitive engagement is the psychological investment in academic achievement and is often demonstrated by a student's ability to conceptualize, organize, and contemplate in the act of deep learning to fully understand the material \cite{chi2014icap}.   

Classical methods for measuring student engagement are self-reports and external behavior observations \cite{Smer2021MultimodalEA}. Self-reports can be used post-lesson or multiple times during the lesson. The former may introduce biases in retrospective recall, while the latter may disrupt the lesson flow. In external behavior observations, student engagement can be evaluated according to the rating instrument, whether using video footage or direct observation in a classroom. Even though external observations are an effective way to measure student engagement, they are not scalable due to their cost and the need for specialized training for human raters. Measuring student engagement in a classroom is important to provide an instructor with a quantitative measure so the instructor can adjust the lecture and classroom activities to re-engage students when disengaged. As a result, disengaged students can be confidentially recognized and assisted. Identifying the engagement of each student in a classroom can be difficult because of the inability of an instructor to focus on all students, especially in classrooms that have a large number of students and free-form seating classrooms in which an instructor cannot observe all students at the same time. Automation of this process would improve the learning process. 

Since 2016, our team has focused on researching students’ engagement using non-intrusive means that do not 
overwhelm the class setting or constitute a burden on students or 
their teacher~\cite{alkabbany2019measuring,farag2021toward, alkabbany2023experimental}. In this work, we focus on automatically measuring behavioral engagement through student body movement. Students’ actions can indicate whether a student is on- or off-task during the lecture. Usually, an engaged student takes notes, types on a laptop, and raises hands more frequently than a disengaged student who performs off-task actions such as playing with mobile, eating/drinking, and checking time.
Hence, the recognition of student actions is an essential step toward the classification of student's behavioral engagement. We have to highlight that classrooms should be a friendly environment in which students can do actions such as drink, check time, etc., and they are still considered to be engaged in the lecture. However, such actions can be done with much less frequency than disengaged students do. Therefore, we depend on the histogram of actions that encodes how frequently a student does an action to classify the student's engagement.

\begin{figure}[t]
\includegraphics[width= 1\linewidth]{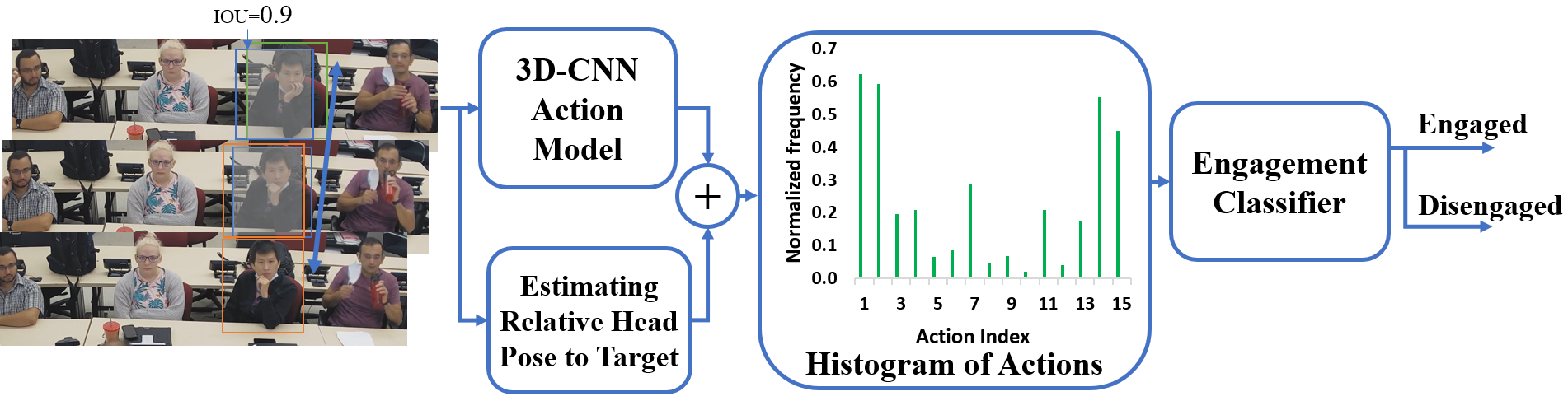}
\caption{The proposed framework for measuring student engagement. First, a 3D-CNN model is used to classify actions in each 2-minute video segment. Also, head poses relative to the target (e.g., blackboard) are estimated. Then the histogram of actions is computed. Finally, a classifier determines the student engagement level based on the extracted histogram of actions. IoU denotes intersection over union.}
\vspace{-15pt}
\label{fig:framework}
\end{figure}

In this paper, we address the following research question: Are the behavioral cues of student's body actions and student's gaze sufficient to estimate student engagement without facial emotions?
To this end, we propose a generic learning framework, see Fig.~\ref{fig:framework}, for measuring behavioral engagement. The proposed framework has three main steps. In the first step, students' skeletons are estimated using the bottom-up pose estimation approach~\cite{cao2017realtime}, which estimates 2D human poses by first predicting all body keypoints in an image and then associating these keypoints to individual people, in each video frame. In the second step, the sequence of the upper body of a student's skeleton is used to generate a volume of 3D pseudo heatmaps by stacking the skeleton joints heatmaps along the temporal dimension. To classify these 3D heatmaps into their corresponding actions, we adopt a 3D-CNN model. Then the trained model is used to generate a histogram of actions from each 2-minute video segment. The calculated histogram of actions encodes the actions and their occurrences. Since certain actions, such as crossing arms, cannot be interpreted without the student's gaze, the frequency of the student looking at the target, e.g., blackboard from such a video segment is calculated using the student's gaze. This frequency is added to the histogram of action which is used as a feature vector to identify student behavioral engagement.
Another problem we address in this work is the lack of a public standard benchmark for 
dataset of students in classrooms. Most of the researchers validate their approaches using private datasets, e.g.,~\cite{zaletelj2017predicting,sun2021student}. Thus, collecting a benchmark dataset for student action recognition and measuring engagement is a necessity for evaluating approaches to student engagement classification. This is one of the objectives of our proposed work. Thus, to evaluate the proposed approach, a dataset is collected and annotated by educational experts.
The contributions of this work include: 
\begin{itemize}
 \item Create a dataset for students in classrooms. The dataset consists of two subsets: i) annotated videos for students performing a set of actions; ii) lectures videos that are annotated by Psychology and Education experts into two engagement levels.
 \item Perform transfer learning for PoseConv3D model~\cite{Duan_poseconv3d} to detect student's actions in a video segment using only the upper body of the student's skeleton.
    \item Develop an approach based on a histogram of actions to identify student's behavioral engagement.
    \item Synthesize disengaged features using kernel density estimate to balance the training data.
\end{itemize}

\begin{table*}
\caption{Summary of student engagement research in the classroom setting.
SVR:linear Support Vector Regression, SVM: Support Vector Machine, RF: Random forest, KNN: K-nearest neighbor, LR: Logistic regression, MLP: Multi-layer perceptron, LSTM: Long short-term memory, BT: Bagged Trees.}
\vspace{-10pt} 
\label{table:summary_classroom_setting}
\begin{center}
 {\footnotesize 
\begin{tabular}{|p{2.5cm}|p{3cm}|p{2cm}|p{1cm}|p{5cm}|}
 \hline
  \multicolumn{1}{|c|}{Study} & \multicolumn{1}{c|}{Visual cues} & \multicolumn{1}{c|}{Models} & \multicolumn{1}{c|}{Interval} & \multicolumn{1}{c|}{Limitations}\\ \hline
 {\footnotesize Janez and Andrej, 2017 \cite{zaletelj2017predicting}} & {\footnotesize head pose, gaze, facial expressions, body posture} &{\footnotesize BT, KNN}&{\footnotesize 1-s}&{\footnotesize each Kinect sensor is used to observe $\leq 4$ students. It is not scalable for large classrooms} \\ \hline
 {\footnotesize Thomas and Jayagopi, 2017 \cite{thomas2017predicting}} &{\footnotesize FACS presence, headpose and eye gaze} & {\footnotesize SVM, LR}&{\footnotesize 10-s}& {\footnotesize students were seating in an arbitrary way that might not accurately depict the circumstances in a real classroom}\\ \hline
 {\footnotesize Goldberg et al., 2021 \cite{goldberg2021attentive}}&{\footnotesize head pose, gaze, FACS action unit intensities} &{\footnotesize SVR}&{\footnotesize 1-s}&{\footnotesize student body movements is not included in estimating engagement even though they annotated their data based on it.}\\ \hline

{\footnotesize {\"O}mer et al., 2021 \cite{Smer2021MultimodalEA}}&{\footnotesize head pose and facial expressions} &{\footnotesize SVM, RF, MLP, and LSTM} &{\footnotesize 1-s}&{ \footnotesize using ResNet-50 network to get attention embedding is dependent on the classroom setting.}\\ \hline
{\footnotesize Mohammadreza and Safabakhsh, \cite{mohammadreza2021lecture}}&{\footnotesize sequence of student's face}&{\footnotesize LBP-TOP, CNN, LSTM, 3D-CNN} &{\footnotesize 10-s} &{ \footnotesize the authors considered student's gaze but not the student's actions in annotating student's engagement.}\\ \hline
{\footnotesize Pabba et al., 2022 \cite{pabba2022intelligent}}&{\footnotesize facial expressions} &{\footnotesize CNN}&-&{ \footnotesize behavioral cues aren't considered in estimating engagement.}\\ \hline

\end{tabular}
}
\end{center}
\vspace{-20pt}
\end{table*}

\section{Related Work}

\textbf{Action recognition} is an active research topic in computer vision literature~\cite{kong2022human} due to its challenges, which include intra- and inter-class variations, camera motion, dynamic background, and viewpoint variations. 
The modalities that are used to recognize human actions include appearance, depth, optical flow, and human skeleton. 

The accuracy of student actions recognition, utilizing both the student's appearance and movements, is highly affected by visual similarity between various actions and the incapacity of optical flow to capture minor student movements~\cite{sun2021student}, e.g., Sun et al.~\cite{sun2021student} reported 61\% accuracy. On the other hand, a skeleton sequence, which encodes both the student's posture and motion, is robust to variations in background and illuminations. Therefore, we adopt this modality for the proposed student's behaviour recognition approach.

\textbf{Skeleton-based action recognition} approaches can be grouped into four categories based on the used network architecture: Graph Convolutional Network (GCN), Convolutional Neural Network (2D-CNN), 3D-CNN, or Recurrent neural network (RNN). In GCN-based approaches \cite{liu2020disentangling}, the sequence of the human skeleton is modeled as spatiotemporal graphs. The limitations of these approaches are non-robustness to noises in pose estimation and the necessity of careful design in integrating the skeleton with other modalities. In 2D-CNN based methods 
\cite{caetano2019skelemotion}, manually designed transformations are utilized to model the sequence of skeletons as a pseudo image. Such input limits the exploitation of the convolutional networks' locality feature, making these techniques less competitive than GCN-based techniques on widely used benchmarks. The RNN-based approach \cite{du2015hierarchical} recognizes simple actions that involve the movement of only a specific human skeleton part and complex actions that involve the movement of a set of body parts through modeling the motion of each body part and their combination. 

In the 3D-CNN based method \cite{Duan_poseconv3d}, the input is represented as heatmaps volume, which captures the structure of skeleton joints and its dynamics over time.  Such input is a holistic representation of an action.
Lin et al.~\cite{lin2021student} recognized student actions in the classroom based on upper body skeleton joints. The drawback of their method is the inability to model the action dynamics due to recognizing student's actions from a single image instead of a video segment. We adopted the 3D-CNN based approach to recognize student actions from a video segment because it can learn spatiotemporal dynamics and is resilient to pose estimate noise through representing the human skeleton as pseudo heatmaps.

\textbf{Measuring student engagement} piques the curiosity of many researchers recently.
Most of the prior research focuses on the computer-based learning setting where reliable visual features of a single student can be extracted \cite{bosch2016using,alkabbany2019measuring}, while few studies addressed the classroom setting, which has challenges of low-resolution visual features, occlusions, and different classroom layouts~\cite{zaletelj2017predicting,yang2020classroom,Smer2021MultimodalEA,alkabbany2023experimental}. Other classroom analytics systems, e.g., EduSense~\cite{ahuja2019edusense,ahuja2021classroom}, simulate the classroom with its objects and occupants and provide a set of audio and visual features.  These systems have not, however, provided evidence of students' engagement.
Table \ref{table:summary_classroom_setting} summarizes recent student engagement studies in a classroom and their drawbacks. 

Janez and Andrej \cite{zaletelj2017predicting} estimated student engagement through a set of features derived from Kinect sensor data. Although their classifier achieved a moderate engagement classification accuracy of 75.3\%, it is not scalable due to Kinect distance limitation.
Thomas and Jayagopi \cite{thomas2017predicting} used mean and standard deviation, within a 10-second video segment, of the student's eye gaze, head rotation, and facial action units to determine the student's engagement level using an SVM classifier. However, students were seated in a particular way and were watching motivating videos, which might not accurately depict the circumstances in a genuine classroom. Ashwin and Guddeti \cite{ashwin2019unobtrusive} classified student's classroom engagement into four categories: not engaged, nominally engaged, engaged in task, and very engaged, based on students' body posture, facial expressions and hand gestures. An improved detection model based on  Single Shot Multi-box detector (SSD) is proposed to detect students and estimate their engagement level. However, the limitation is that student engagement is estimated from a single frame, which may not accurately reflect true engagement levels that requires a proper time resolution. 

Goldberg et al. \cite{goldberg2021attentive} utilized a continuous observer-based rating system that integrates ICAP framework~\cite{chi2014icap} with on-/off-task behavior analysis. The authors found a substantial correlation between manual annotations of engagement and students' self-reports as well as knowledge tests. They used an SVM regressor, whose estimate is correlated with the students' self-reports.
 {\"O}mer et al. \cite{Smer2021MultimodalEA} determined the engagement level of secondary school students from facial videos in a classroom using feature embedding for attentional and affective features. However, the attentional embedding features are skewed towards the classroom environment in use, and any changes to this setting will require additional model training. Also, their dataset annotation~\cite{goldberg2021attentive,Smer2021MultimodalEA}  relies on student actions, but they only used facial features for the engagement model. 

Mohammadreza and Safabakhsh~\cite{mohammadreza2021lecture} investigated four approaches that extract spatiotemporal information in a 10$s$ facial video to gauge students' attention. Extracted features using each model were used to train a separate neural network for student's attention classification. Their best result was a 0.82 F1-score using the 3D-CNN model. They used the student's gaze and head movement in annotation, but not the student's actions which limit their method's ability to handle the multi-dimensional nature of the engagement.
Pabba et al. \cite{pabba2022intelligent} developed a CNN model to recognize a student's affective state from a student's facial image. The engagement level of a student is determined by the student's affective states without taking behavioral cues into account. Pabba and Kumar \cite{pabba2023vision} utilized student's facial expressions, head poses, and head movements to classify student's classroom engagement into engaged and non-engaged. Their approach  classifies student engagement level based on a weighted summation of three scores: affect score from positive and negative academic affective states, attention score from frontal and non-frontal head poses, and head movement score based on head displacement distance.
Although their approach obtained an accuracy of 0.75, it is validated using self-reported engagement levels, which may not accurately reflect the actual engagement.
Three studies in~\cite{pabba2022intelligent,mohammadreza2021lecture,pabba2023vision} assessed student's engagement through student's facial expressions or gazes while ignoring their actions. Even though a student writes a note, they may still be deemed disengaged based on these approaches.

\section{Proposed Framework}
\label{sec:framework}
Three main steps comprise the proposed framework (see Fig.~\ref{fig:framework}): i) bottom-up student pose estimation; ii) student action recognition; and iii) student on-/off-task classification. 
The students' body joints are first detected in each frame using the bottom-up pose estimation approach~\cite{cao2017realtime}, and then we associate each student's pose using the intersection over union (IoU) criterion. 
 Second, for each 2-minute  video segment, we apply a sliding time window technique to extract a set of frames from which heatmaps, representing the upper body joints, are generated and fed into a 3D-CNN network for action classification. Finally, a histogram of actions is calculated, from the identified actions in the 2-minute time interval, and then that histogram with the student’s gaze is used by an engagement classifier to determine the student's engagement level. Since the dataset is collected in real classes where students’ behaviors spontaneously occur, most students are typically engaged, whether they are taking notes, folding their arms, or putting their hands on their faces while looking at the target. To address the issue of the limited number of disengaged samples and the imbalance in students’ actions, we employ data synthesis, and weighted cross-entropy loss, respectively.

\subsection{Actions Dictionary}

\begin{figure}[t]
\includegraphics[width=1\textwidth]{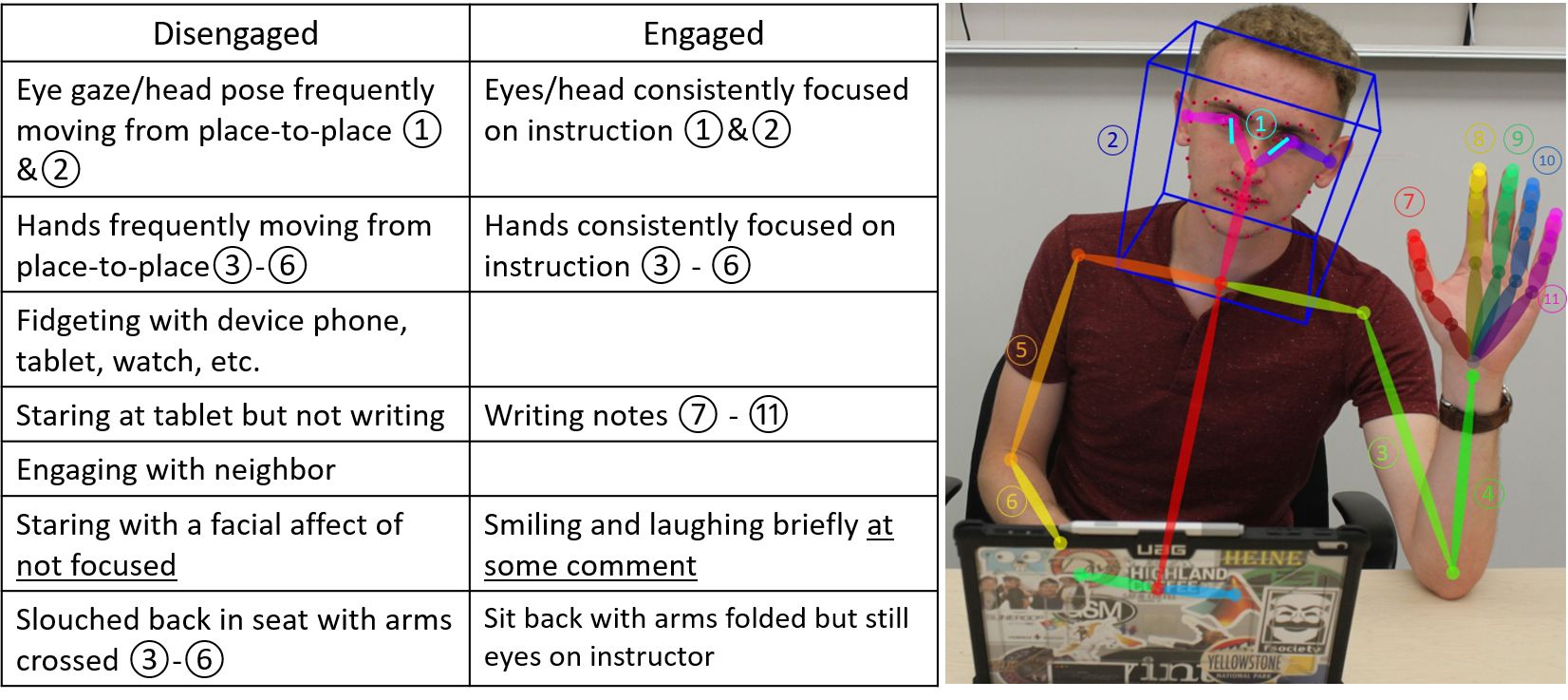}
\caption{Dictionary of the proposed tokens.}
\label{fig:dictionary}
\end{figure}

\begin{figure}[t]
\includegraphics[width=1\textwidth]{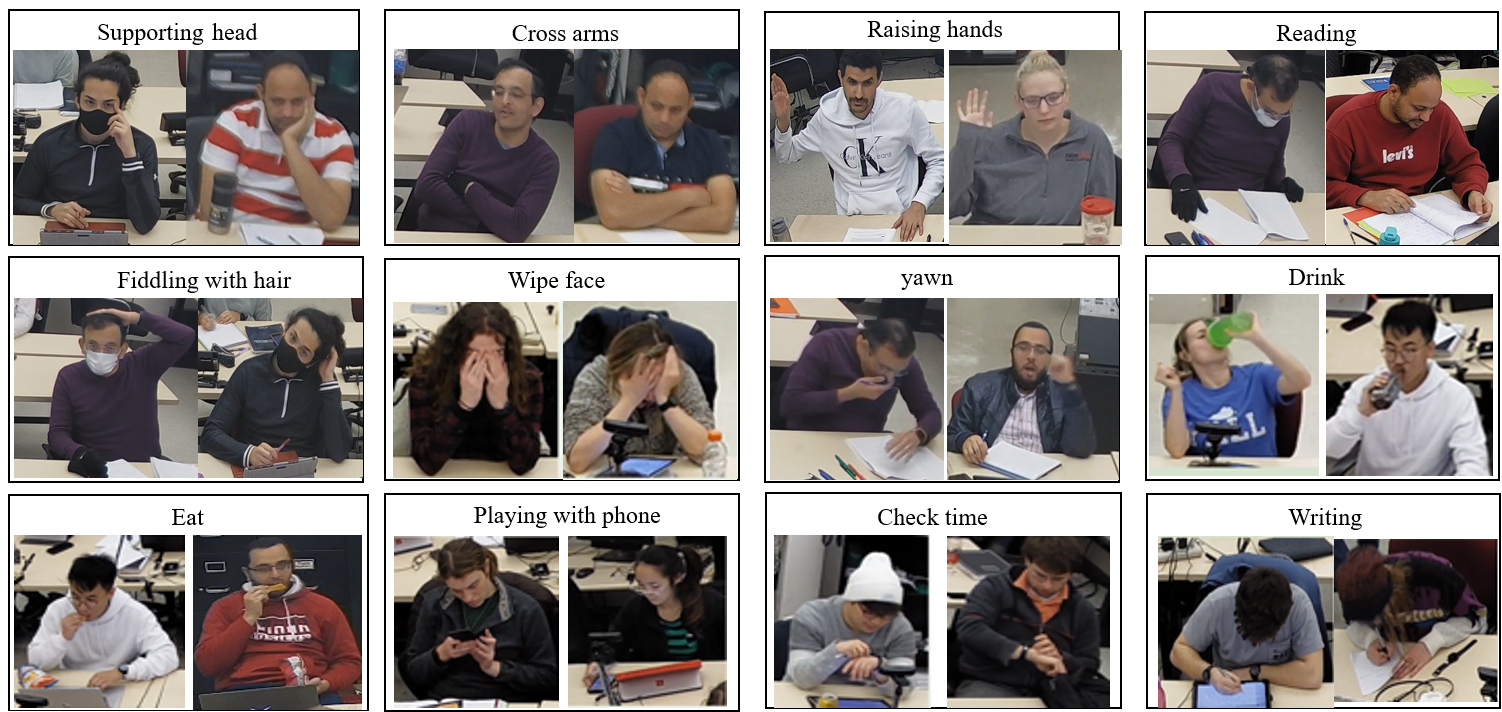}
\caption{Examples of student's actions in a classroom.}
\label{fig:dataset_examples}
\vspace{-10pt}
\end{figure}

The student's behavioral engagement can be estimated from their actions during the lecture. We conducted a pilot study where Psychology and Education experts watched a set of recorded lectures in a classroom. 
In this study, each 2-minute video was coded using two engagement levels: disengaged and engaged. Each engagement level can be described by a set of measurable actions (tokens). A dictionary of these tokens is constructed to identify the observable measures of each engagement level, as described in Fig.~\ref{fig:dictionary}. Based on the conclusion of this study, we define a set of actions that indicate an engaged student, such as writing, raising hands, reading, and typing on a keyboard. We also define a different set of behaviors, like playing with a phone or tablet, wiping your face, yawning, checking the time on a watch, fiddling with your hair, drinking water, and eating snacks, that are indicative of a disengaged student. Additional gestures, like crossing arms and supporting the head, should be used in conjunction with the student gaze rather than being used alone for classification. Examples of students' actions are shown in Fig.~\ref{fig:dataset_examples}. The proposed approach calculates how frequently each action occurs within a 2-minute window and then utilizes that histogram and student’s gaze as a feature vector to determine the level of student engagement.

\subsection{Student Actions Recognition}

To recognize student's actions, we build on the state-of-the-art skeleton-based approach~\cite{Duan_poseconv3d}, in which a sequence of human skeletons is represented as a 3D pseudo heatmap volume, which will be described below, and then a 3D-CNN is used for classification. 

\begin{figure}[t]
\centering
\includegraphics[scale=.9]{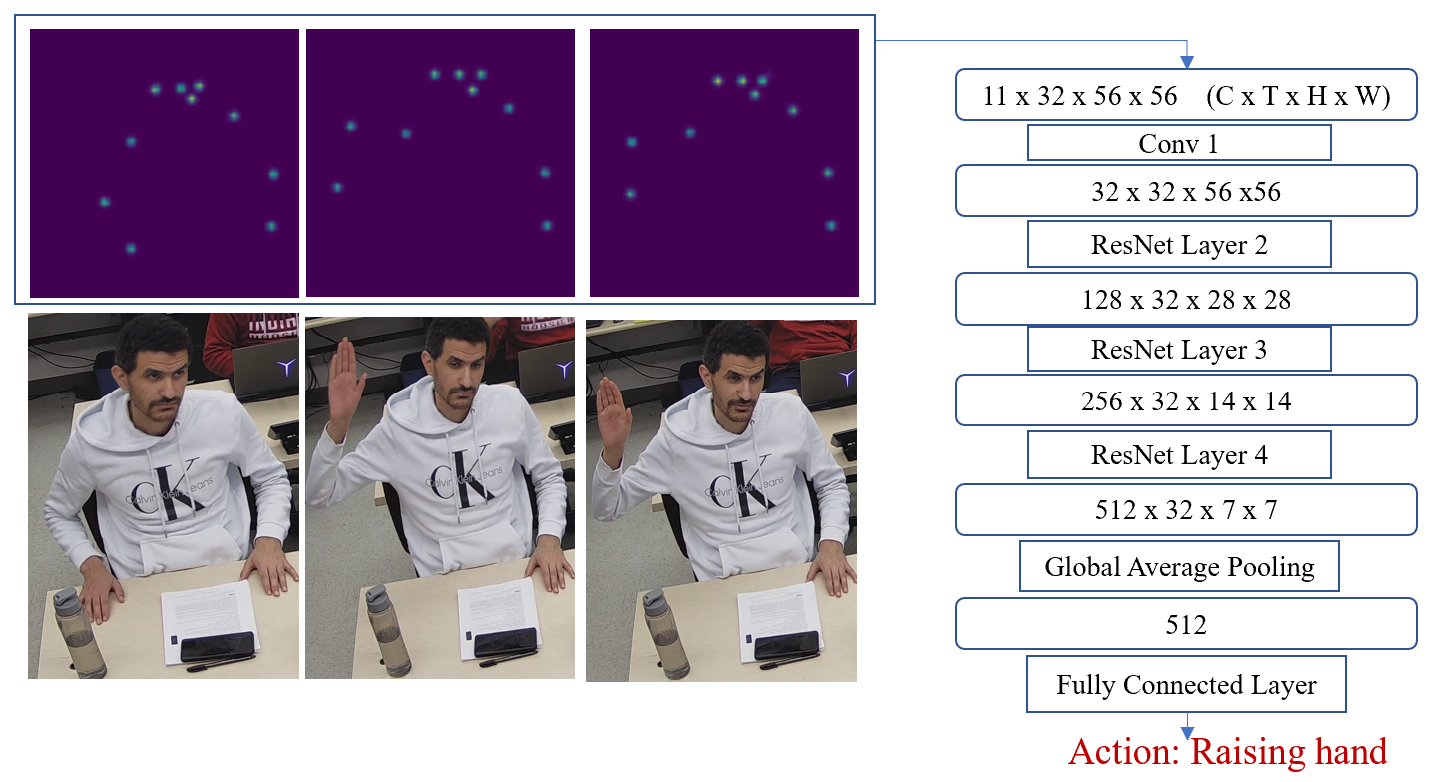}
\caption{The 3D-CNN architecture for action recognition. Given a student video clip, skeleton joints in each frame are extracted, and then a volume of 3D pseudo heatmaps is generated by stacking the upper body of student skeleton joints heatmaps along the temporal dimension. Finally, such volume is used as an input for the 3D-CNN model that consists of 5 layers to recognize the containing action.}
\label{fig:framework2}
\end{figure}

We propose two modifications to the architecture in~\cite{Duan_poseconv3d}: i) a weighted cross-entropy loss function is used to address the class imbalance presented in our dataset, and ii) only the joints in the upper body are used because the desks that students sit behind occlude other joints.  
Lastly, due to the limited size of our dataset, we perform transfer learning by training the model (Fig. ~\ref{fig:framework2}) using upper body joints on the NTU-60 dataset~\cite{shahroudy2016ntu}, freezing the first two layers of the model, and then training the final three layers. 

\textbf{3D pseudo heatmaps.} Given coordinate-triplets ($x_k,y_k,c_k)$ of skeleton joints for each frame, the heatmaps of K joints are generated using gaussian map centered at each joint:
\begin{equation} H_k(x,y) = \exp{- \frac{(x-x_k)^2 + (y-y_k)^2}{2\sigma^2}}\end{equation}
where $x_k$and $y_k$ are the coordinates of $k$th joint and $\sigma$ controls Gaussian map variance.
After that, the K heatmaps of all frames are stacked along the temporal dimension to form the 3D heatmap volume with size $K\mathrm{x}T\mathrm{x}H\mathrm{x}W$ where K is a number of human body keypoints, T is temporal length, and H and W are the height and width of such maps.

\subsection{Student's Gaze Estimation}
\label{studentGaze}
Estimating student's eye gaze in the classroom from wall-mounted cameras is a difficult task. 
Head orientation, as previously demonstrated in studies \cite{raca2013system,ahuja2019edusense} serves as a reliable surrogate for gaze attention in classrooms.
To estimate the student's head pose in different setup configurations of classrooms, first, student facial landmarks are estimated using \cite{Prados-Torreblanca_2022_BMVC}. Second, head pose is estimated using Perspective-n-Point (PnP)-based method that finds correspondence between estimated facial keypoints on an image and their corresponding 3D locations in an anthropological face model. 
Finding head pose is used to identify if a student looks at the target or not through finding the intersection point between the gaze direction vector and the target.
 
\subsection{Student Engagement Estimation}
\begin{figure}
    \centering
    \begin{subfigure}[b]{0.45\textwidth}
         \centering
         \includegraphics[width=\textwidth]{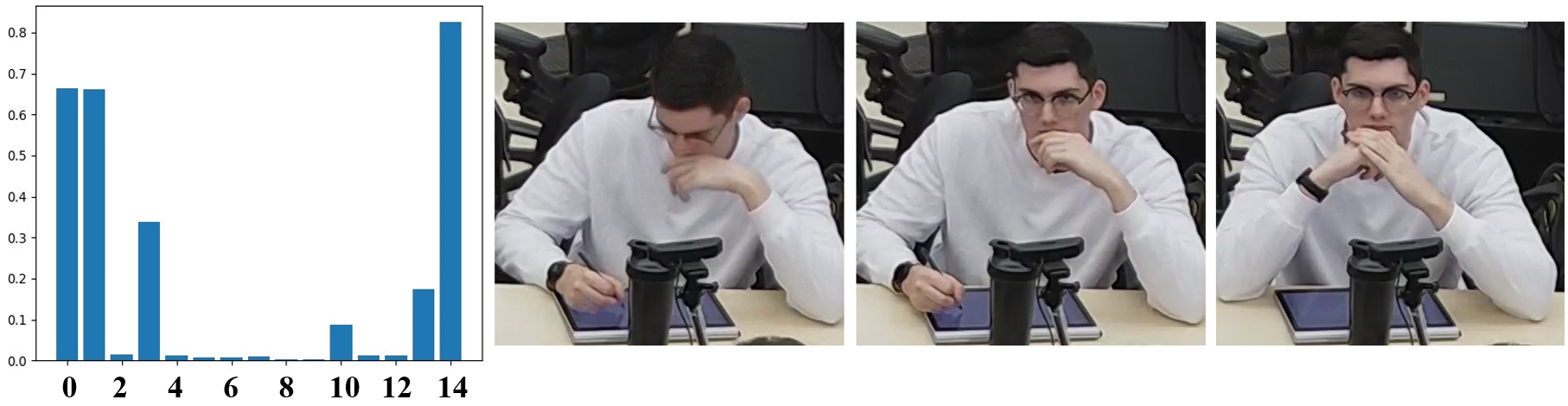}
         \caption{Engaged sample}
         \label{fig:f_engaged}
     \end{subfigure}
     \hfill
     \begin{subfigure}[b]{0.45\textwidth}
         \centering
         \includegraphics[width=\textwidth]{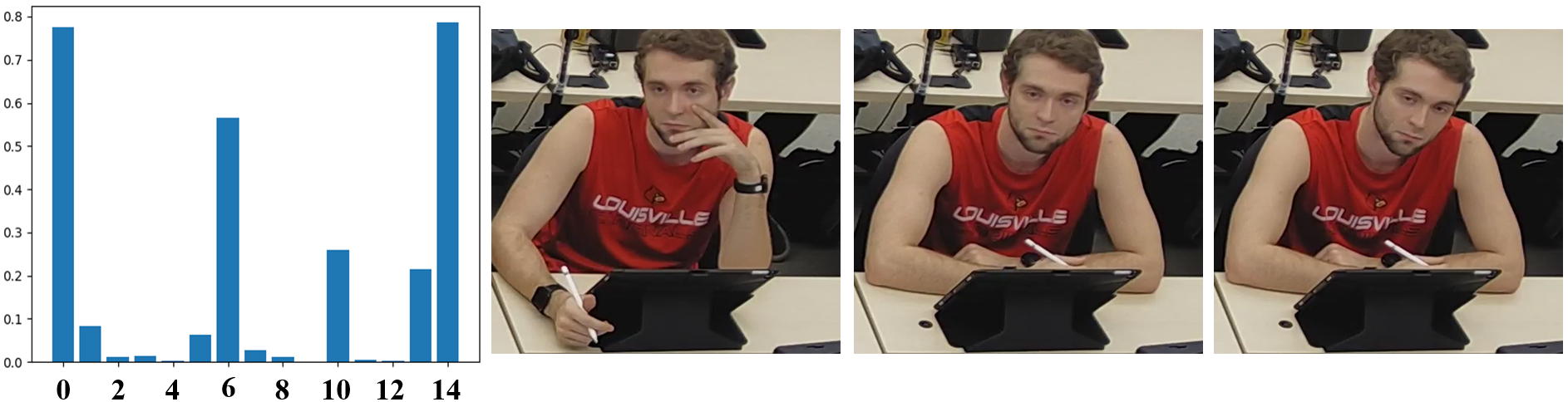}
         \caption{Engaged sample}
         \label{fig:s_engaged}
     \end{subfigure}
     \hfill
     \begin{subfigure}[b]{0.45\textwidth}
         \centering
         \includegraphics[width=\textwidth]{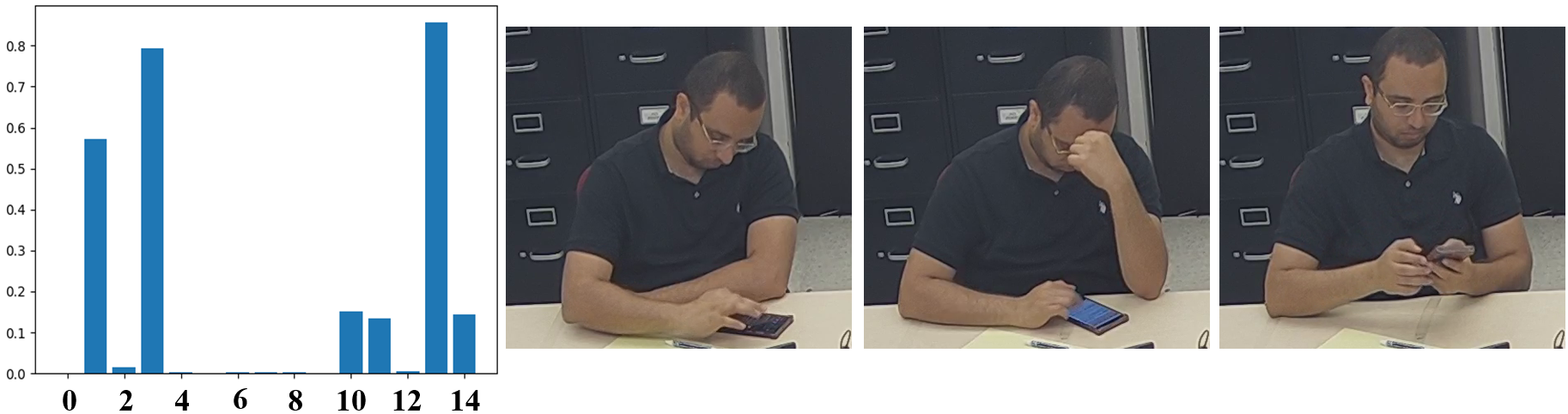}
         \caption{Disengaged sample}
         \label{fig:f_disengaged}
     \end{subfigure}
     \hfill
     \begin{subfigure}[b]{0.45\textwidth}
         \centering
         \includegraphics[width=\textwidth]{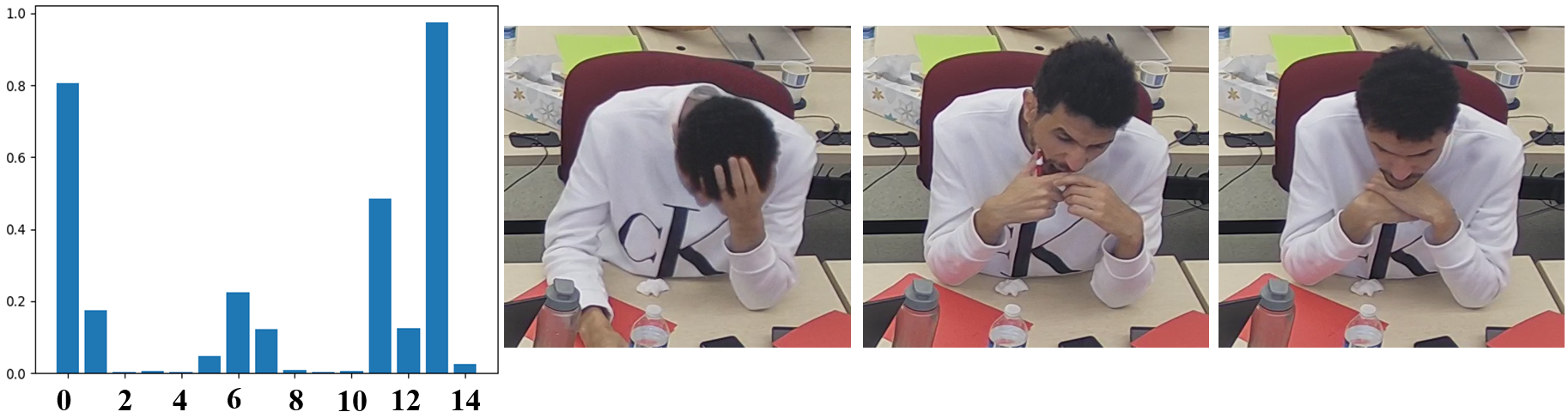}
         \caption{Disengaged sample}
         \label{fig:s_disengaged}
     \end{subfigure}
    \caption{Examples of student behavior engagement features. (a) and (b) are engaged samples while (c) and (d) are disengaged samples. x-axis represents the action index and y-axis represents the normalized frequency of such action. The actions are 0: supporting head, 1: writing, 2: typing on a keyboard, 3: playing with phone/tablet, 4: reading, 5: raising hands, 6: cross arms, 7: wipe face, 8: drink water, 9: eat meal/snack, 10: check time, 11: fiddling with hair, 12: yawn 13:look off target, 14:look at target.}
    \vspace{-15pt}
    \label{fig:featuresExamples}
\end{figure}

To determine whether a student is engaged (on-task) or disengaged (off-task), we use the histogram of actions, because it not only encodes the presence of the actions but also represents how frequently the student performs an action. As shown in Fig.~\ref{fig:dictionary}, experts interpret an action based on its frequency. On the other hand, actions such as crossing arms and supporting head cannot be interpreted without the student's gaze. So, we estimate the student's head pose relative to a target as described in section \ref{studentGaze}. Then we calculate how frequently the student looks at the target during the 2-minute interval. Finally, the normalized frequencies of actions and gazes represent the feature vector, which is fed into a classifier to identify whether the student is engaged or disengaged. Examples of such feature vectors are illustrated in Fig.~\ref{fig:featuresExamples}. 

Furthermore, when student behavior is shown as a histogram of actions, model predictions are easier to interpret. As an example, Fig.~\ref{fig:f_engaged} shows an engaged student who spends the majority of the time staring at the target, placing a hand on the face, and taking notes. Another illustration of an engaged student who looks at the target, supports his head and crosses arms is provided in Fig.~\ref{fig:s_engaged}.
 On the other hand, Fig.~\ref{fig:s_disengaged} represents a disengaged student who looks off the target while supporting head. This confirms the importance of the student's gaze when interpreting an action. Another illustration for a disengaged student who plays with phone while supporting head is shown in Fig.~\ref{fig:f_disengaged}. 

\subsection{Histogram Synthesizing}
\label{subsec:Histogram_Synthesizing}
Due to the small number of disengaged samples (64 disengaged samples in training compared to 577 engaged samples) in our student engagement dataset, we propose a method for synthesizing new samples that more accurately reflect the characteristics of disengaged samples.
The synthesizing process is described as follows: 
\begin{enumerate} 
\item  Given the disengaged samples of the training set $\{x_1,x_2,...,x_n\}$ where $x_i \in \mathbb{R}^{15}$ is the i-th disengaged sample.
\item kernel density estimate (KDE) is employed to estimate $\hat{f}_X$ at a new sample x as follows: \(\hat{f}_X(x) = \frac{1}{n} \sum_{i=1}^{n}  \phi_\sigma(\frac{|x-x_i|}{\sigma})  \) where $\phi_\sigma $ is the Gaussian density with mean zero and standard-deviation $\sigma$. In other words, the density estimate at each sample is the average contribution from each kernel at that sample. 
\end{enumerate}

\vspace{-20pt}
\section{Experiments}
\label{sec:Experimental_results}
\subsection{Student behavioral Engagement Dataset}
\label{sec:cvip15_dataset}
\begin{figure}[t]
\centering
\includegraphics[width=0.6\textwidth]{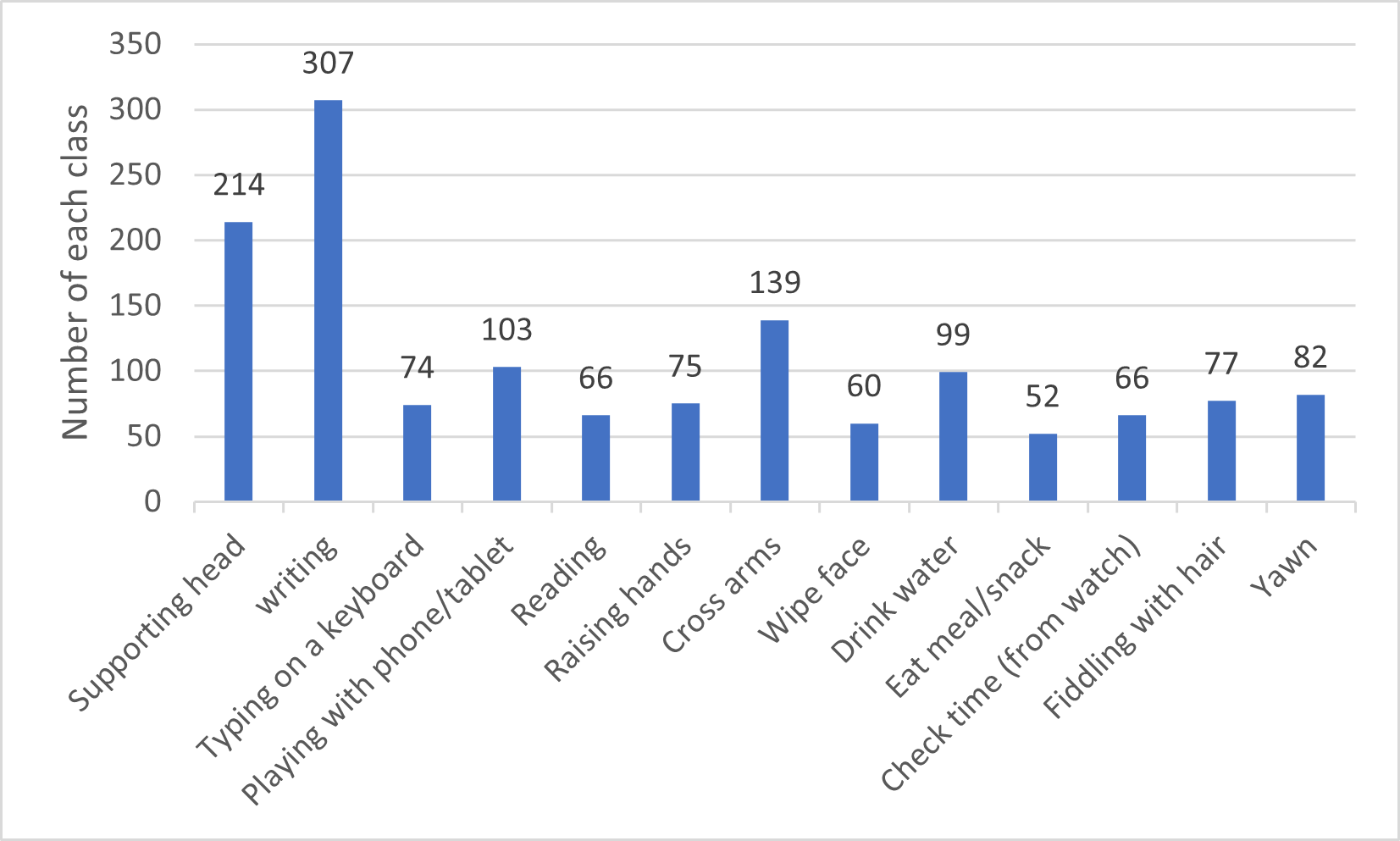}
\caption{Statistics of student actions dataset where the number of video clips per each class is shown.}
\vspace{-20pt}
\label{fig:dataset_stats}
\end{figure}

In order to assess the efficacy of the proposed approach, we gathered a behavioral engagement dataset comprising two subcategories: student actions and student engagement. This is required because neither student engagement nor classroom behavior has a public standard benchmark. The videos of the dataset were captured using a wall-mounted camera in a classroom with a 4K resolution and 15 fps. The sizes of classes range from 6 to 11 students. For student actions, we annotated 1414 video segments, which were recorded from 4 courses. Also, 13 videos, one for each action, were recorded in a controlled setting. The statistics of the annotated videos are illustrated in Fig.~\ref{fig:dataset_stats}.  
Videos from four one-hour lectures for eleven students make up the student engagement dataset. Experts in education and psychology annotated these videos. This annotation divides each two-minute video segment into engaged and disengaged categories.   

\vspace{-10pt}

\subsection{Implementation Details}
In our implementation, we trained the PoseConv3D model on NTU-60 dataset \cite{shahroudy2016ntu} using upper body joints for left and right arms, left and right eyes and ears, as well as the nose. After that, we performed fine-tuning for the model weights to train it on student actions by freezing the first two layers of the model's backbone. The fine-tuning process involves 140 epochs with a batch size of 8 and a learning rate of 0.0025.

\subsection{Experimental Results}
The proposed approach for action recognition is evaluated by dividing the student action dataset into training and test sets with a ratio of 3 to 1 for each action. 
The top 1 and mean accuracy of our action recognition model are $86.32\%$ and $85.48\%$, respectively. The top 1 and mean accuracy are not very high due to the overlap among many actions. There are subtle differences between different action categories, such as writing and supporting head, where there is a variant in the writing class in which students put their hands on their faces during writing. 

To evaluate the behavioral engagement classifier, the four lectures are split into three lectures for training and one lecture for testing. The training set contains 577 engaged samples and 64 disengaged samples. The test set contains  277 engaged samples and 45 disengaged samples. To overcome the imbalance in the dataset, we performed undersampling of the engaged class using the Edited Nearest Neighbor (ENN) method~\cite{Wilson4309137} with L2 distance metric in all experiments. The ENN method discards samples that have a majority of opposite-class neighbors, identifying them as potential noise or ambiguous samples. The engaged samples in the training set after undersampling are 462. In addition, we generated synthetic samples using the proposed synthesizing method described in subsection \ref{subsec:Histogram_Synthesizing}. Empirically, the number of generated disengaged samples is set to 20. Hence, the disengaged samples in the training set after synthesis are 84. 
Table.~\ref{table:Evaluation_metric} shows the recall, precision, and F1-score of the random forest engagement classifier. We notice that the F1-score of the random forest classifier that did not use synthetic samples in training is 92\% and 60\% for engaged and disengaged classes, respectively. The reason for the inferior F1-score of disengaged is the high imbalance in the dataset. However, the proposed framework obtains a moderate weighted average f1-score. On the other hand, we can notice that the random forest classifier that used synthetic samples in training has better performance than its counterpart that did not use these samples. This indicates that the generated disengaged samples help in reducing the effect of the data imbalance.
Another reason for the inferior F1-score of disengaged is that, as shown in Figure ~\ref{fig:dictionary} both behavioral and emotional cues are used to annotate student engagement. 
Thus, taking emotional cues into account improves the F1-score of disengaged class 2\% as described in subsection \ref{sec:behavioral_emotional_cues}.

\begin{figure}[t]
   \centering
\includegraphics[scale=0.8]{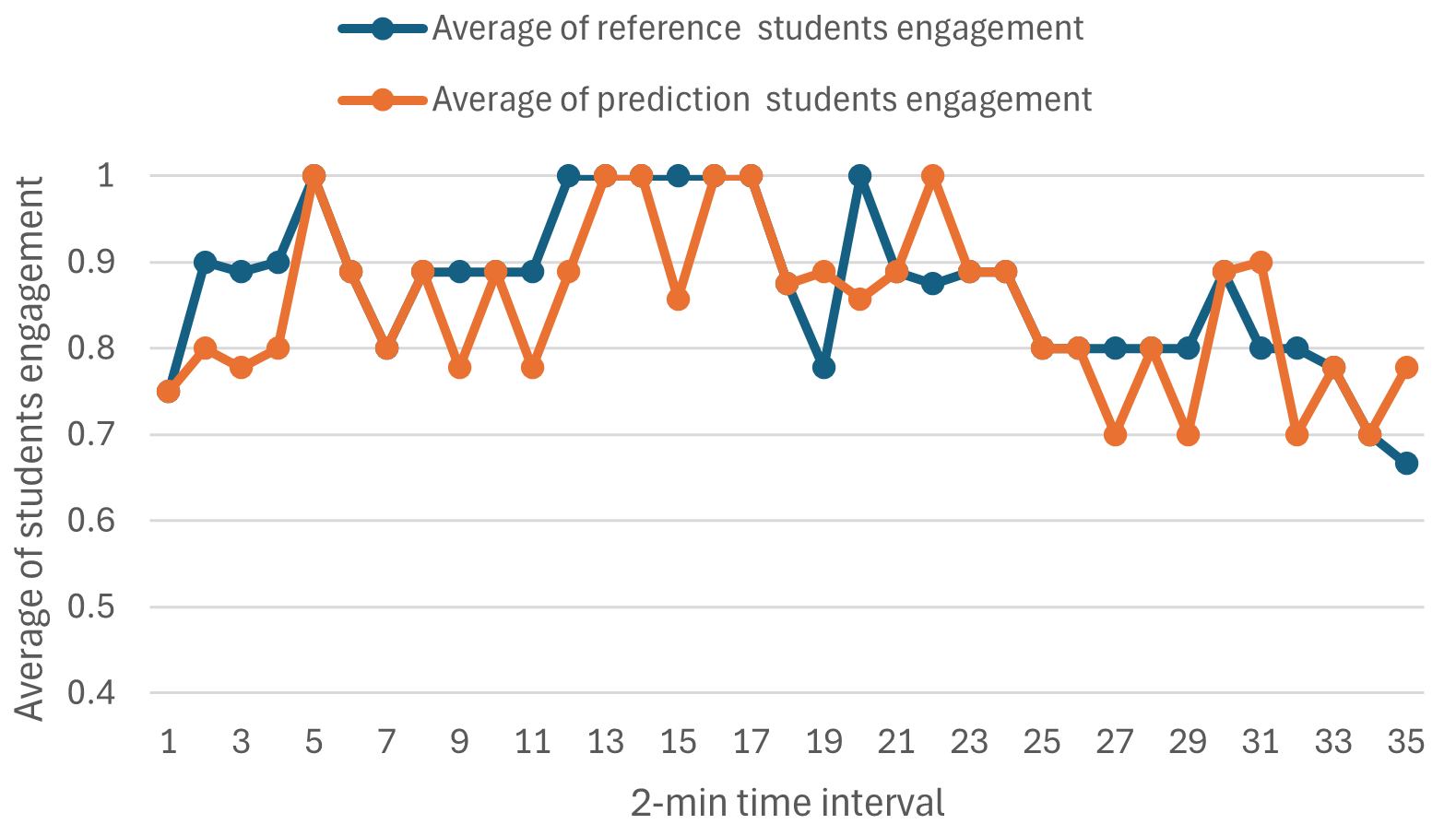}
\caption{Average engagement of 10 students during the testing lecture. The mean of reference engagement level is displayed as a blue line, while the mean of prediction engagement level is illustrated as an orange line. The study was conducted on 10 students, however, at each 2-minute period, the engagement level may be averaged over different numbers of students due to a student’s nonexistence (e.g., stepping out) at such a time period or a student was completely occluded. }
\vspace{-15pt}
\label{fig:mean_comparison}
\end{figure}

A comparison between the ground truth of the average students' engagement and the average of predicted student engagement levels for the test lecture is shown in Fig.~\ref{fig:mean_comparison}. There is a correlation between the prediction and the ground truth, which confirms that although the model does not include emotional cues, it can still capture the average engagement of the class.

\subsection{Comparison to State-of-the-art}
Although, it is challenging to directly compare our measurements of student engagement with those of other studies, due to differences in datasets and annotation techniques as shown in Table \ref{table:comparison_with_others}, we apply the algorithms in~\cite{thomas2017predicting,mohammadreza2021lecture} on our dataset to estimate student engagement. We implemented the two algorithms by extracting the necessary features from our dataset and using the best-performing classifier reported in each approach. To extract the required features, we calculated the mean and the standard deviation of head rotation per 10-second for \cite{thomas2017predicting} and extracted a sequence of 75 students facial regions from 10-second video segments for \cite{mohammadreza2021lecture}.
The annotation in \cite{thomas2017predicting,mohammadreza2021lecture} depends on whether the student is looking at the target or not. So, we annotated our dataset accordingly with the help of the student's head pose.  

For comparison purposes, since our experts' annotation was done every 2-minute,  we took the majority vote on the predictions of these algorithms in a 2-minute window. Table \ref{table:comparison_ours_with_comp} shows the comparison of our method with these competitors. It is evident that our proposed method achieved better performance than the compared methods.
The performance of the first approach~\cite{thomas2017predicting}, on our dataset, is 45\% F1-score, while the reported F1-score, on their dataset, is 89\%. Also, the method in  \cite{mohammadreza2021lecture} achieved 30\%  and 82\% F1-scores on our dataset and their dataset, respectively.
These discrepancies in F1 scores are because the authors determined a student's engagement based on whether the student looked at the target within 10 seconds or not, but they neglected to consider other aspects of engagement such as students' actions.

 
\begin{table*}
\caption{Comparison of proposed method with state-of-the-art student engagement estimation methods.}
\vspace{-10pt}

\label{table:comparison_with_others}
\begin{center}
 {\footnotesize 
\begin{tabular}{|p{3cm}|p{1.0cm}|p{2cm}|p{3.5cm}|p{2.75cm}|p{2cm}|}
 \hline
  \multicolumn{1}{|c|}{Study} & \multicolumn{1}{c|}{\makecell {Device}} & \multicolumn{1}{c|}{\makecell {Data Annotaion}} & \multicolumn{1}{c|}{Classifier's input}&\multicolumn{1}{c|}{\makecell {Classifier}}  & \multicolumn{1}{c|}{Metric}\\ \hline
 {\footnotesize Janez \& Andrej \cite{zaletelj2017predicting}} &{\footnotesize Kinect} & {\footnotesize 7320 1-s videos} & {\footnotesize  median filter over 1-s frames' features } &{\footnotesize KNN}  & {\footnotesize Accuracy: 0.75}  \\ \hline
 {\footnotesize Thomas \& Jayagopi \cite{thomas2017predicting}}& 
 {\footnotesize Camera}  &{\footnotesize 2263 10-s videos} &{\footnotesize frames' feature aggregation} & {\footnotesize SVM}  &{\footnotesize AUC: 0.81}  \\ \hline
 {\footnotesize Ashwin \& Guddeti \cite{ashwin2019unobtrusive}  } & {\footnotesize Camera}&{\footnotesize 4423 frames}&{ \footnotesize single frame}  &{\footnotesize scale-invariant 2D CNN}   &{ \footnotesize AUC: 0.70}   \\ \hline

{\footnotesize Mohammadreza \& Safabakhsh \cite{mohammadreza2021lecture}}&{\footnotesize Camera}&{\footnotesize 6500 10-s videos}&{ \footnotesize  10-s videos}  &{\footnotesize 3D-CNN}   &{ \footnotesize  F1-score: 0.82}   \\ \hline
{\footnotesize Pabba \& Kumar \cite{pabba2023vision}} & {\footnotesize Camera} &{\footnotesize 450 2-min videos} &{\footnotesize frames' affect, attention and head movement scores aggregation} & {\footnotesize weighted summation of three scores} & {\footnotesize Accuracy: 0.75} \\ \hline
{\footnotesize Proposed method}&{\footnotesize Camera}&{\footnotesize 963 2-min videos}&{ \footnotesize histogram of student's actions and gaze} &{\footnotesize Random forest} & \footnotesize {F1-score: 0.90}  \\ \hline
\end{tabular}
}
\end{center}
\vspace{-30pt}
\end{table*}
\begin{table}
\caption{Evaluation metric of random forest student engagement classifier.}
\vspace{-15pt}
\label{table:Evaluation_metric}
\begin{center}
 {\footnotesize
\begin{tabular}{|c|c|c|c|c|} 
 \hline
   {\footnotesize Synthesizing}&&{\footnotesize Recall} &{\footnotesize  Precision} &{\footnotesize  F1-score} \\ \hline
\multirow{3}{1em}{{\footnotesize without}}
    &{\footnotesize disengaged} &{\footnotesize 0.69}  &{\footnotesize 0.53}   &{\footnotesize 0.60}    \\ 
   &{\footnotesize engaged }&{\footnotesize 0.90}  &{\footnotesize    0.95 }  & {\footnotesize  0.92}    \\\cline{2-5} 
   &{\footnotesize weighted avg}&{\footnotesize 0.87}  & {\footnotesize   0.89 }  & {\footnotesize  0.88 }   \\ \hline

   \multirow{3}{1em}{{\footnotesize with}}
   &{\footnotesize disengaged }&{\footnotesize  0.67} &{\footnotesize 0.61} &{\footnotesize 0.64 }      \\ 
   &{\footnotesize engaged }&{\footnotesize    0.93} & {\footnotesize 0.95} &{\footnotesize  0.94}          \\\cline{2-5} 
   &{\footnotesize weighted avg} &{\footnotesize 0.89} &{\footnotesize  0.90 }&{\footnotesize 0.90}\\ \hline
\end{tabular}
}
\end{center}
\end{table}

\begin{table}
\caption{Comparison of our method with competitors.}
\vspace{-15pt}
\label{table:comparison_ours_with_comp}
\begin{center}
{\footnotesize
\begin{tabular}{|c|c|c|c|c|c|} 
 \hline
  \;\;\;\;\;\; Method\;\;\;\;\;\; &&Recall &Precision & F1-score \\ \hline
   \multirow{3}{6em}{Thomas and Jayagopi \cite{thomas2017predicting}}
   &disengaged & 0.96  &    0.19   &   0.32    \\ 
   &engaged & 0.33  &    0.98   &   0.49    \\\cline{2-5} 
   & weighted avg& 0.42 &0.87     & 0.47    \\ \hline

\multirow{3}{6em}{Mohammadreza and Safabakhsh\cite{mohammadreza2021lecture}}
&disengaged & 0.49  &    0.15   &   0.23    \\ 
   &engaged & 0.56  &    0.87   &   0.68    \\\cline{2-5} 
   & weighted avg&0.55& 0.77  &     0.62     \\ \hline

   \multirow{3}{3em}{Ours}&disengaged &  0.67 &0.61 & 0.64       \\ 
   &engaged &    0.93 & 0.95 &  0.94          \\\cline{2-5} 
   &weighted avg & 0.89 & 0.90 & 0.90\\ \hline
\end{tabular}
}
\end{center}
\end{table}
\begin{table}
\caption{Comparison between different classification methods.}
\vspace{-15pt}
\label{table:comparison_diff_methods}
\begin{center}
{\footnotesize 
\begin{tabular}{ |c|c|c|c|c|c} 
 \hline
  Method &&Recall &Precision & F1-score \\ \hline
   \multirow{3}{3em}{Decision tree}
   &disengaged & 0.58  &    0.50   &   0.54    \\ 
   &engaged & 0.91  &    0.93   &   0.92    \\\cline{2-5} 
   & weighted avg& 0.86  &    0.87   &   0.86    \\ \hline

\multirow{3}{3em}{Neural network}
&disengaged & 0.60  &    0.47   &   0.52    \\ 
   &engaged & 0.89  &    0.93   &   0.91    \\\cline{2-5} 
   & weighted avg& 0.85  &    0.87   &   0.86    \\ \hline

   \multirow{3}{3em}{SVM}&disengaged & 0.84  &    0.42   &   0.56    \\ 
   &engaged & 0.81  &    0.97   &   0.88    \\\cline{2-5} 
   & weighted avg& 0.81  &    0.89   &   0.84    \\ \hline

    \multirow{3}{3em}{Random forest} &disengaged & 0.69  &    0.53   &   0.60    \\ 
   &engaged & 0.90  &    0.95   &   0.92    \\\cline{2-5} 
   & weighted avg& 0.87  &    0.89   &   0.88    \\ \hline
\end{tabular}
}
\end{center}
\end{table}
\begin{table}
\caption{Evaluation metric for student engagement using actions and gaze, separately.}
\vspace{-20pt}
\label{table:action_gaze_Evaluation_metric}
\begin{center}
{\footnotesize
\begin{tabular} {|c|c|c|c|c|} 
 \hline
   Features&&Recall &Precision & F1-score \\ \hline
   \multirow{3}{4em}{Frequency of actions}
   &disengaged &  0.69 &0.30 & 0.42       \\ 
   &engaged &    0.74 & 0.94 &  0.83          \\\cline{2-5} 
   &weighted avg & 0.73 & 0.85 & 0.77\\ \hline
 \multirow{3}{4em}{Gaze}
  &disengaged &  0.84 &0.29 & 0.43       \\  
   &engaged &    0.67 & 0.96 &  0.79          \\\cline{2-5}
   &weighted avg & 0.69 & 0.87 & 0.74\\ \hline

\end{tabular}
}
\end{center}
\end{table}

\subsection{Ablation Studies}
In this section, we conducted a comparison between different engagement classifications and then evaluated the best-performing one using student actions and student's gaze, separately. In these experiments, we use the original training data without synthesizing disengaged samples. 

\subsubsection{Comparison of different engagement classifiers}
Table \ref{table:comparison_diff_methods} shows the performance of the decision tree, neural network, random forest, and SVM classifiers on our student engagement subset. The neural network consists of 2 fully connected layers with 32-neuron and 1-neuron with a 1-way sigmoid, respectively. From Table \ref{table:comparison_diff_methods}, we observe that the performance of random forest is better than that of decision tree, neural network and SVM classifier with 2\%, 2\%, and 4\% in F1-score, respectively. Despite the decision tree and neural network having the same weighted average F1-score, the decision tree is good at recognizing disengaged with precision 50\% compared with 47\% of neural network. Random forest classifier is preferred in this case because it provides a good precision-recall trade-off. Hence, such a classifier gives the instructor low false alarms regarding the number of disengaged students while teaching the class, so the lecturer may adjust the lecture and class activities to re-engage these students.

\subsubsection{Student Engagement using Actions and Gaze, separately}
To evaluate the contribution of each feature in determining student engagement level, we compared the performance of the random forest classifier based on features of student actions and student's gaze, separately in Table \ref{table:action_gaze_Evaluation_metric}. We can notice that measuring student engagement using the frequency of student actions is better than using the ratio of looking at the target (Gaze) in terms of the weighted average F1 score. Actions, such as crossing arms and supporting head, still need the student's gaze to determine whether the student is engaged or disengaged. In addition, we can observe the gaze is effective to determine the disengaged students, although students who are writing, reading, and typing on the keyboard may be considered disengaged due to their low rate of looking at the target. Thus, we can conclude that student's actions and gaze are complementary to each other and both are important for measuring engagement level.

\subsubsection{Student Engagement using behavioral and Emotional Cues }
\label{sec:behavioral_emotional_cues}
In this section, we study the effect of including emotional cues representing students' affective states along with behavioral cues representing students' actions and their relative gaze toward the target on the performance of student engagement in the classroom. To assess students' affective states, several works \cite{bosch2015automatic,xu2023spontaneous} suggest using learning-centered affective states (focused, confused, bored, and delighted) instead of basic emotions (anger, fear, sadness, happiness, disgust, and surprise), as the latter are infrequent in the learning context. To this end, we collected a dataset containing 1881 images of students' faces, categorized as follows: focused (1401), confused (100), bored (83), and delighted (297). Three annotators classified these images into four categories, achieving a Fleiss’ Kappa of 0.41, indicating moderate agreement. To balance the dataset, we used 300 samples from the focused category. We then trained EfficientNet \cite{tan2019efficientnet} on AffectNet facial expression dataset \cite{mollahosseini2017affectnet}, performing transfer learning on the collected dataset by freezing all layers of the model except the classification layer. To get the student's engagement level using behavioral and emotional cues, the frequency of the student's affective states during the 2 minutes time interval together with frequency of student actions and the frequency of the student looking at the target are used as inputs to random forest classifier. 
Although adding emotional features enhances the classification of the disengaged class, it does not enhance the classification of the majority class, i.e., engaged one. We believe that this is due to that the most cues that have been used by annotators to measure students engagement are the behavioral cues. This is confirmed by the results in Table \ref{table:action_gaze_emotion_Evaluation_metric}  where the weighted average F1-score of our proposed method (frequency of actions and gaze) with synthetic samples outperforms that of using frequency of actions, gaze and emotions with 1\%.
\begin{table}
\caption{Evaluation metric for student engagement using actions, gaze and emotions.}
\vspace{-20pt}
\label{table:action_gaze_emotion_Evaluation_metric}
\begin{center}
{\footnotesize
\begin{tabular} {|c|c|c|c|c|} 
 \hline
   Features&&Recall &Precision & F1-score \\ \hline
   \multirow{3}{8em}{Frequency of actions, gaze and emotions}
   &disengaged &  0.78 &0.57 & 0.66       \\ 
   &engaged &    0.91 & 0.96 &  0.93          \\\cline{2-5} 
   &weighted avg & 0.89 & 0.91 & 0.89\\ \hline
\end{tabular}
}
\end{center}
\vspace{-25pt}
\end{table}
\vspace{-10pt}
\section{Conclusion and Future Work}
\vspace{-7pt}
In this paper, we presented a novel framework for classifying student engagement using student actions and the relative pose of the student's head to the instructor board. First, a sequence of the upper part of the skeleton for each student is fed into a 3D-CNN model for action recognition. Then, a random forest classifier is utilized to predict the student engagement level based on the histogram of the detected actions and the student’s gaze. The proposed framework is tested on the created dataset and achieves a performance that is sufficient to capture the average engagement of the class. This performance implies that student's actions and their occurrences reveal student's level of engagement inside the classroom.

The future work involves collecting more video clips for student actions to achieve outstanding action recognition performance and overcome the imbalance in the engagement classes. 
\bibliographystyle{elsarticle-num}



\end{document}